\documentclass[runningheads]{llncs}

\usepackage[T1]{fontenc}
\usepackage{graphicx}
\usepackage{amsmath}
\usepackage{amssymb}
\usepackage{color}
\begin{document}
\title{Estimating 2D Keypoints of Surgical Tools
Using Vision-Language Models with Low-Rank
Adaptation}
\titlerunning{Estimating 2D Keypoints of Surgical Tools Using VLM with LoRA}
% If the paper title is too long for the running head, you can set
% an abbreviated paper title here
%
\author{Krit Duangprom \and
Tryphon Lambrou \and
Binod Bhattarai}
\authorrunning{K. Duangprom et al.}
%\authorrunning{Anonymous et al.}
% First names are abbreviated in the running head.
% If there are more than two authors, 'et al.' is used.
\institute{University of Aberdeen, Aberdeen, UK\\
\email{\{k.duangprom.24, tryphon.lambrou, binod.bhattarai\}@abdn.ac.uk}}
\maketitle              % typeset the header of the contribution
\begin{abstract}
This paper presents a novel pipeline for 2D keypoint estimation of surgical tools by leveraging Vision Language Models (VLMs) fine-tuned using a low rank adjusting (LoRA) technique. Unlike traditional Convolutional Neural Network (CNN) or Transformer-based approaches, which often suffer from overfitting in small-scale medical datasets, our method harnesses the generalization capabilities of pre-trained VLMs. We carefully design prompts to create an instruction-tuning dataset and use them to align visual features with semantic keypoint descriptions. Experimental results show that with only two epochs of fine tuning, the adapted VLM outperforms the baseline models, demonstrating the effectiveness of LoRA in low-resource scenarios. This approach not only improves keypoint detection performance, but also paves the way for future work in 3D surgical hands and tools pose estimation. 
% Further optimization of the low-rank tuning strategy is expected to improve accuracy and expand applicability to more complex surgical environments.

\keywords{Keypoint Estimation  \and Vision-Language Models \and Low-Rank Adaptation.}
\end{abstract}
\section{Introduction}

Understanding surgical tools is essential for the advancement of intelligent medical imaging, where accurate 2D keypoint estimation underpins crucial tasks such as 3D pose estimation~\cite{wu2025surgpose}, spatial analysis~\cite{hamza2025automated}, and immersive AR/VR-based surgical simulations~\cite{hein2025next}. These keypoints not only enable recognition of the geometry and orientation of the object, but also facilitate precise downstream tasks such as motion tracking and simulation-based surgical training~\cite{bkheet2023using}. Despite its importance, annotated data for surgical tool landmarks remains scarce, hindering progress in this area. If we can reliably transform 2D observations into 3D spatial understanding, we can gain deeper insights into surgical performance, such as identifying procedural mistakes, analyzing tool usage patterns, and offering real-time guidance for training or even in-theater decision support.

Current approaches for surgical tool keypoint detection typically rely on Convolutional Neural Networks (CNN) or Vision-Transformers~\cite{wu2025surgpose,spektor2024monocular,aboukhadra2024surgeonet} trained on task-specific datasets. Although these models have shown strong performance in constrained setups, they often suffer from overfitting and lack generalizability, especially when applied to complex or novel tool types. Moreover, retraining such models for each specific task limits the scalability.

To address these limitations, we propose a new pipeline that leverages large Vision-Language Models (VLMs)~\cite{bai2025qwen2,wu2024deepseek} for estimating 2D keypoints of surgical tools. Using the generalization ability of pre-trained VLMs in diverse image-text pairs, our method can be adapted to the surgical domain using Low-Rank Adaptation (LoRA)~\cite{hu2022lora}—a lightweight fine-tuning technique that injects task-specific knowledge without requiring full retraining of the base model.

Our work introduces a novel application of Vision Language Models (VLMs) for surgical tool landmark detection using lightweight adaptation, demonstrating that these models can perform structured localization tasks beyond open-ended image understanding. This approach lays the groundwork for extending VLM-based frameworks to multi-object scenes involving hand-tool interactions and paves the way toward full 3D ~\cite{wang2023pov,dong2024hamba,qi2024hoisdf} pose estimation in real surgical environments. Looking ahead, this pipeline has the potential to support keypoint estimation for both hands and tools, forming a foundation for more complex vision tasks such as dynamic interaction modeling and semantic scene interpretation in surgical workflows.

\section{Related Work}
\noindent \textbf{Convolutional Neural Network (CNN)-Based Approaches:}
CNN-based methods are fundamental in keypoint and pose estimation, including applications such as surgical tool landmark detection. These models take RGB images as input and are trained to predict landmarks of tool parts.

The typical pipeline uses annotated datasets where the network learns to regress keypoint coordinates or heatmaps. Loss functions such as Mean Squared Error (MSE) are used to compare predictions with ground truth, guiding the model to improve localization accuracy. Among CNN-based methods, High-Resolution Network (HRNet)~\cite{sun2019deep} is a benchmark for precise 2D keypoint detection, maintaining high resolution features throughout the network. This is especially useful in tasks requiring fine-grained localization, such as hand or face landmarks. Alternatively, YOLO-Pose~\cite{maji2022yolo} offers a single-stage approach that jointly predicts object bounding boxes and keypoints.

\noindent \textbf{Transformer-Based Approaches:}
Recent work applies ViT backbones to pose estimation by combining global and fine-grained features. In surgical tool tracking, Transformer-based models outperform CNNs in handling occlusions and intricate structures. SurgeoNet~\cite{aboukhadra2024surgeonet} improved YOLO-Pose by adding transformer layers and using stereo cameras to improve keypoint detection and 7D pose estimation.

\noindent \textbf{Vision-Language Models:}
Vision-Language Models (VLMs) aim to unify visual and textual information for semantic understanding~\cite{shrestha2023}. A foundational model in this space is CLIP (Contrastive Language–Image Pretraining)~\cite{radford2021learning}, which learns a joint embedding space for images and text using contrastive loss. CLIP comprises two encoders: a visual encoder (e.g., ViT) that transforms images into feature vectors and a text encoder (e.g., transformer) that embeds language prompts.

Advanced VLMs such as Qwen-VL 2.5~\cite{bai2025qwen2} and DeepSeek-VL 2~\cite{wu2024deepseek} introduce instruction tuning and enhance visual grounding. Qwen-VL 2.5 supports fine-grained spatial reasoning and conversational prompting, while DeepSeek-VL 2 is optimized for accurate localization in open-ended tasks. Both models can respond to detailed prompts about object parts or pose, offering strong potential in fields like medical imaging and robotic perception.

\noindent \textbf{Pose and Keypoint Estimation with Vision-Language Models:}
Subramanian et al.~\cite{subramanian2025pose} propose a novel approach where pose optimization is guided by language descriptions. They convert LLM-generated textual cues such as body parts interactions or spatial constraints into training losses, enabling zero-shot pose supervision without dense keypoint labels.

Other studies have demonstrated that VLMs can distinguish between different tool components. For example, Huang et al.~\cite{huang2024combining} leverage language-grounded visual representations to segment tool parts (e.g., handles). Similarly, Gong et al.~\cite{gong2024zerokey} show that VLMs can be prompted to reason about object geometry and estimate approximate keypoint positions, such as corners or edges, without requiring task-specific fine-tuning.

To adapt VLMs more directly for keypoint and pose estimation, fine-tuning techniques have been introduced. CLIPose~\cite{lin2024clipose} improved CLIP by incorporating point cloud data with images and text, improving more accurate pose prediction. Recently, KPT-LLM~\cite{yang2024kptllm} and CapE-LLM~\cite{kim2024capellm} leverage LoRA~\cite{hu2022lora} lightweight fine-tuning of VLM to localize 2D landmarks such as joints in animals based on prompt-driven supervision.

These approaches demonstrate the growing potential of VLMs to support pose and keypoint tasks through semantic grounding, few-shot generalization, and adaptable supervision.

\section{Method}

We propose a vision-language-based pipeline for semantic 2D keypoint estimation of surgical tools. We reformulate the task as a prompt-based Visual Question Answering (VQA) problem using a large Vision-Language Model (VLM). The model takes as input an RGB image and a fixed natural language prompt:

\begin{quote}
\textit{``What is/are this/these tool(s) and find 12 keypoints?''}
\end{quote}

\begin{figure}[htbp]
\centering
\includegraphics[width=\textwidth]{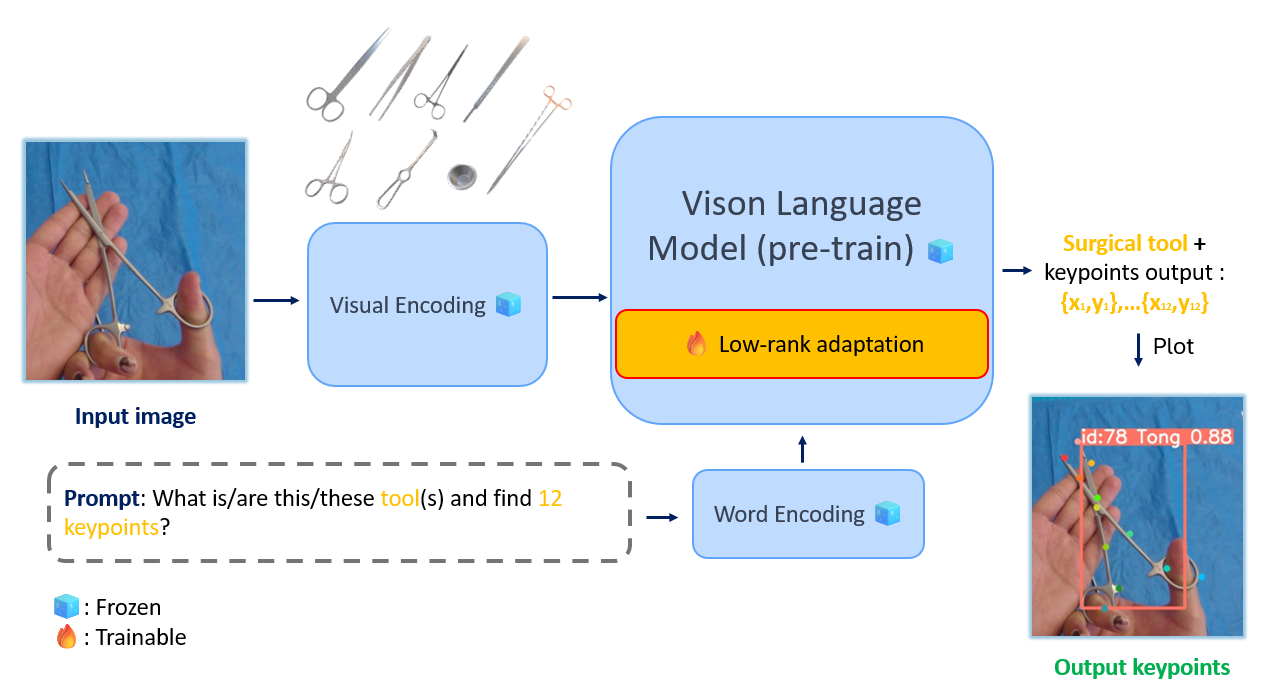}
\caption{Overview of the proposed VLM-based keypoint estimation pipeline. The model receives an image and a prompt, and outputs the tool name along with its 12 semantic keypoints.}
\label{fig:idea_VLM}
\end{figure}

Given this input, the model is trained to directly predict the tool name and the $(x, y)$ coordinates of 12 semantic keypoints. An overview of the proposed pipeline is illustrated in Figure~\ref{fig:idea_VLM}. As shown in the figure, the VLM processes both the image and the prompt jointly. It attempts to describe visual content and identify the tool(s) present in the scene, while also predicting the $(x, y)$ coordinates of 12 semantic keypoints associated with the tool. The output is a text sequence that includes the tool name followed by the predicted keypoints in coordinate format.

After generating the output, the predicted keypoints are parsed and compared with the corresponding ground-truth annotations from the training dataset. The learning objective is to minimize the difference between the predicted and true output sequences. To achieve this, we use the standard causal language modeling (CLM) loss, which guides the model to generate each token in the output sequence autoregressively:

\begin{equation}
\mathcal{L}_{\mathrm{LM}} = -\frac{1}{T} \sum_{t=1}^{T} \log p_\theta\left(y_t \mid y_{<t}, \text{image}, \text{prompt} \right)
\end{equation}

Here, $y_t$ represents the $t$-th token in the target sequence of length $T$, and $p_\theta$ is the predicted token distribution. The sequence includes both the tool name and the numeric keypoint coordinates. This loss encourages the model to learn both semantic understanding and spatial reasoning through text generation.

\subsection*{Fine-Tuning with LoRA}

We apply Low-Rank Adaptation (LoRA)~\cite{hu2022lora} in MS-swift~\cite{zhao2025swift} for efficient fine-tuning by injecting trainable low-rank updates into frozen transformer weights. Specifically, a frozen weight $\mathbf{W}_0 \in \mathbb{R}^{d \times h}$ (e.g. in self-attention) is augmented as:

\begin{equation}
\mathbf{W} = \mathbf{W}_0 + \mathbf{A}\mathbf{B}, \quad \mathbf{A} \in \mathbb{R}^{d \times r},\ \mathbf{B} \in \mathbb{R}^{r \times h}
\end{equation}

\begin{figure}[htbp]
\centering
\includegraphics[width=0.8\textwidth]{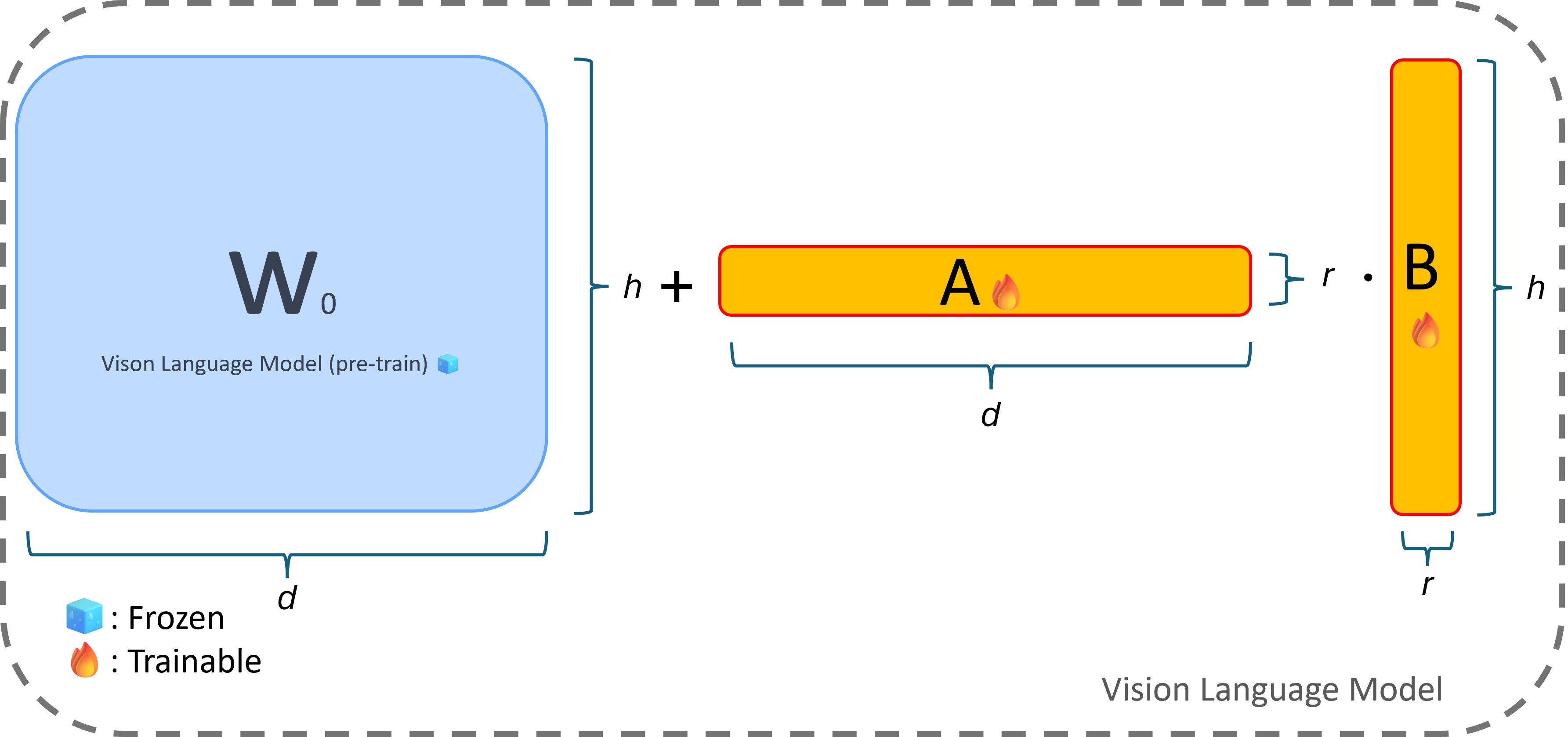}
\caption{Visualization of Low-Rank Adaptation (LoRA): A frozen weight matrix is augmented with a low-rank update via two trainable matrices $\mathbf{A}$ and $\mathbf{B}$.}
\label{fig:idea_LoRA}
\end{figure}

Only $\mathbf{A}$ and $\mathbf{B}$ are trained; $\mathbf{W}_0$ remains frozen. We use rank $r = 8$, scaling factor $\alpha = 16$, and dropout 0.05. LoRA is applied to all attention projections (query, key, value, output) and optionally to feed-forward layers. The weight is illustrated in Figure~\ref{fig:idea_LoRA}

\section{Experiments}
\begin{figure}[htbp]
\centering
\includegraphics[width=\textwidth]{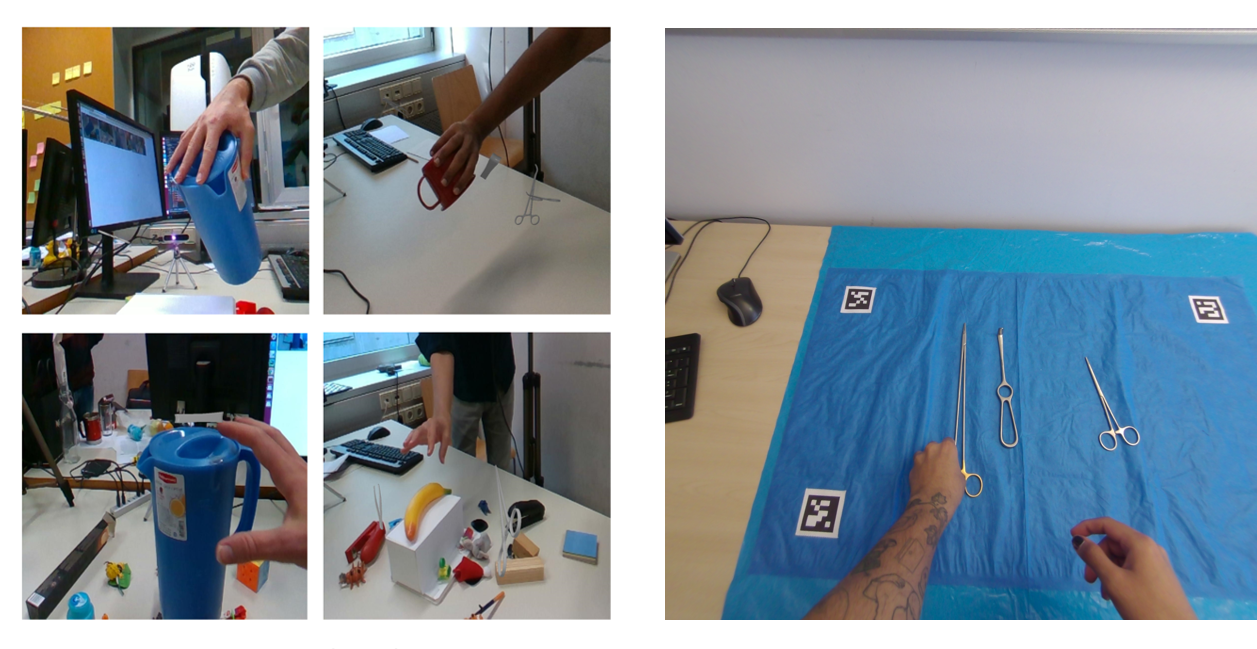}
\caption{Examples from the SurgeoNet dataset: synthetic images (left) and real-world captures (right).}
\label{fig:surgeonet_examples}
\end{figure}

\noindent \textbf{Dataset:}
We performed our experiments SurgeoNet dataset~\cite{aboukhadra2024surgeonet} which has 13,649 training and 1,763 test RGB images covering 14 instrument types. Each image contains one or more surgical tools annotated with 12 semantic 2D keypoints. Example images are shown in Figure~\ref{fig:surgeonet_examples}.

\noindent \textbf{Evaluation Metrics:} We evaluate keypoints estimation using two standard metrics. Mean Per Joint Position Error (MPJPE) measures the average Euclidean distance between predicted and ground truth keypoints as shown below:
\begin{equation}
\text{MPJPE} = \frac{1}{N} \sum_{i=1}^{N} \left\| \hat{\mathbf{K}}_i - \mathbf{K}_i \right\|_2
\end{equation}

Similarly, Percentage of Correct Keypoints (PCK) computes the fraction of keypoints within a normalized distance threshold $\alpha$.
\begin{equation}
\text{PCK}(\alpha) = \frac{1}{N} \sum_{i=1}^{N} \mathbb{I} \left( \frac{\left\| \hat{\mathbf{K}}_i - \mathbf{K}_i \right\|_2}{L} < \alpha \right)
\label{ref:eqn_pck}
\end{equation}
In Equation~\ref{ref:eqn_pck}, $L$ is a normalization factor (e.g., image width), and $\mathbb{I}$ is the indicator function.

% \section{Results}
\subsection*{Experimental Results}

We compare our proposed pipeline with strong baselines for 2D keypoint estimation, including YOLOv8-Pose~\cite{yolov8_ultralytics}, a CNN-based  with 26.4M parameters, and SurgeoNet~\cite{aboukhadra2024surgeonet}, which adds a ViT backbone (1M extra parameters) on top of YOLOv8m. The baselines were fully trained for more than 100 epochs. In contrast, our method uses large vision-language models including Qwen2.5-VL-3B and DeepSeek-VL2-tiny—fine-tuned via LoRA for prompt-based keypoint prediction. Despite training for only two epochs, they achieve competitive results while updating a small fraction of parameters: 20.7M of 3.78B for Qwen, and 38.8M of 3.41B for DeepSeek. We evaluated the perforamnce using MPJPE and PCK at thresholds 0.05 and 0.10. The results are shown in Table~\ref{tab:vlm_results}.

\begin{table}[htbp]
\centering
\caption{Comparison of models on MPJPE and PCK metrics. Lower MPJPE is better; higher PCK is better.}
\resizebox{\textwidth}{!}{
\begin{tabular}{l|c|c|c|c}
\hline
\textbf{Model} & \textbf{MPJPE} $\downarrow$ & \textbf{PCK@0.05} $\uparrow$ & \textbf{PCK@0.10} $\uparrow$ & \textbf{Trainable Params} $\downarrow$ \\
\hline
YOLOv8-Pose & 0.0672 & 0.6572 & 0.8466 & 26.4 M\\
SurgeoNet (monocular camera) & 0.0651 & 0.6711 & 0.8519 & 1 M (on top of Yolov8) \\
\hline 
DeepSeek-VL2 (w/o fine-tune) & 0.4251 & 0.0076 & 0.0296 & - \\ 
DeepSeek-VL2 (fine-tune)     & 0.0796 & 0.6432 & 0.7961 & 38.8 M\\
\hline 
Qwen2.5-VL-3B (w/o fine-tune) & 0.4270 & 0.0067 & 0.0300 & - \\
Qwen2.5-VL-3B (fine-tune)     & \textbf{0.0627} & \textbf{0.6767} & \textbf{0.8908} & \textbf{20.7 M} \\
\hline
\end{tabular}
}
\label{tab:vlm_results}
\end{table}

\subsection*{Ablation Study on LoRA Rank}
We conducted an ablation study to examine the impact of LoRA rank when fine-tuning Qwen2.5-VL-3B, fixing the scaling factor to 16 and varying the rank among $\{4, 8, 16\}$. Results are shown in Table~\ref{tab:ablation_lora_rank}. Rank 8 achieved the best overall performance. A lower rank (4) significantly degraded performance indicating insufficient capacity to adapt the pretrained model effectively. Increasing the rank to 16 slightly improved over rank 4 but still fell short of rank 8, suggesting that overly high ranks may lead to suboptimal convergence or mild overfitting under the same training conditions.

\begin{table}[htbp]
\centering
\caption{Ablation study of LoRA rank on Qwen2.5-VL-3B.}
\resizebox{0.7\textwidth}{!}{
\begin{tabular}{l|c|c|c}
\hline
\textbf{LoRA Rank} & \textbf{MPJPE} $\downarrow$ & \textbf{PCK@0.05} $\uparrow$ & \textbf{PCK@0.10} $\uparrow$ \\
\hline
Rank = 4 & 0.0955 & 56.25\% & 79.81\% \\
Rank = 8 & \textbf{0.0627} & \textbf{67.67\%} & \textbf{89.08\%} \\
Rank = 16 & 0.0796 & 62.42\% & 85.00\% \\
\hline
\end{tabular}
}
\label{tab:ablation_lora_rank}
\end{table}

\subsection*{Visualization of Results}

\begin{figure}[h]
\centering
\includegraphics[width=\textwidth]{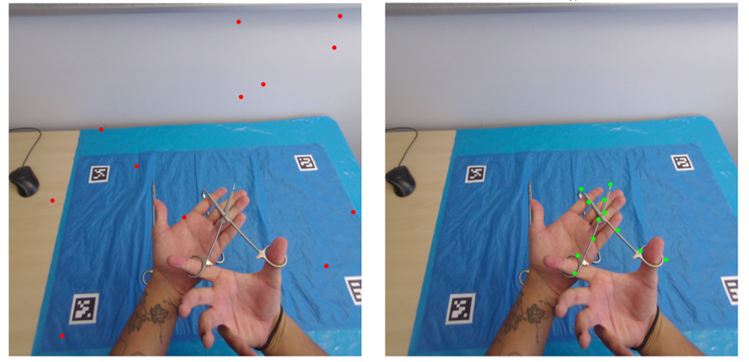}
\caption{Keypoint predictions before (left) and after (right) LoRA fine-tuning. Example shows real surgical images with annotated tool keypoints.}
\label{fig:vlm_vis}
\end{figure}

\begin{figure}[h]
\centering
\includegraphics[width=\textwidth]{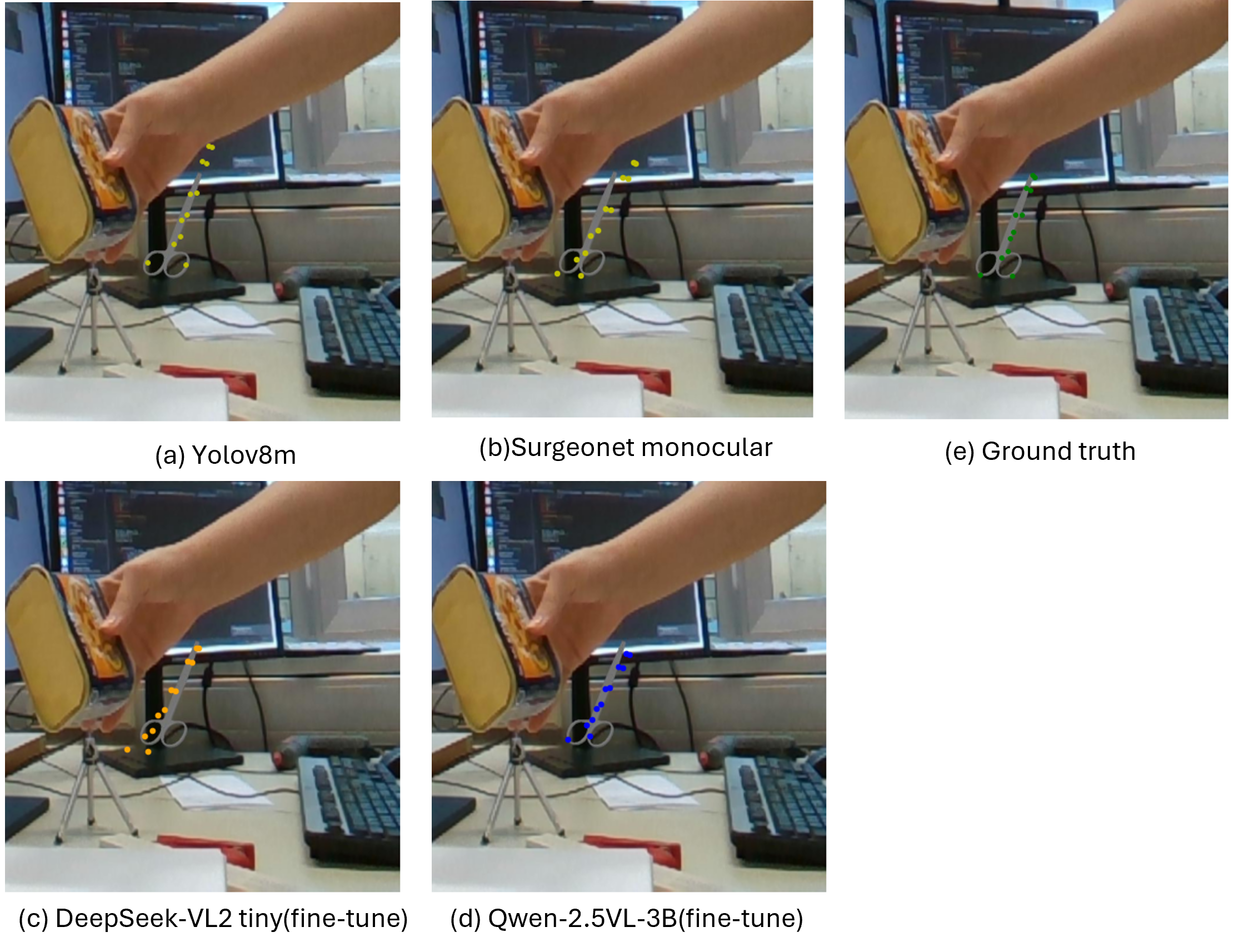}
\caption{Comparison keypoints prediction of models in synthetic test data}
\label{fig:vlm_vis2}
\end{figure}

\section{Discussion}

Our results demonstrate that Vision-Language Models (VLMs) can perform 2D keypoint estimation at a level comparable to CNN or Transformer-based vision models. As shown in Figure~\ref{fig:vlm_vis2}, the VLM-based predictions are qualitatively similar to those from specialized architectures such as YOLOv8-Pose and SurgeoNet. However, Qwen2.5-VL shows a slightly better fit to these scaling challenges, producing more stable and accurate predictions.

Before fine-tuning, the VLMs did not produce meaningful keypoint predictions, as they were not originally trained for this task. However, after applying LoRA-based fine-tuning, the models successfully localize all 12 keypoints aligned with the surgical tool structure (see Figure~\ref{fig:vlm_vis}).

While making comparisons across different large Vision Language Models, we observed that Qwen2.5-VL outperforms DeepSeek-VL2 despite having fewer trainable parameters during fine-tuning. This suggests that vision language pretraining can have a significant impact on downstream performance, even in highly structured spatial tasks with keypoint estimation.

We also studied the effect of the LoRA rank on Qwen2.5-VL-3B, as the rank determines the additional capacity available for task adaptation. We observed that using too small of a rank limited the model’s ability to learn spatially complex relationships, leading to underfitting, while using an overly high rank did not provide additional benefits and even reduced stability during training. A moderate rank provided the best balance, allowing the model to capture detailed geometric patterns while maintaining convergence efficiency. This highlights that careful selection of LoRA rank is essential when adapting large vision-language models to dense spatial prediction tasks.

\section{Conclusions and Future Works}
In this work, we proposed a Vision Language Model (VLM) pipeline for estimating 2D keypoints of 14 types of surgical tools using Low-Rank Adaptation (LoRA) fine-tuning. We designed tailored prompts and reconstructed a synthetic dataset to effectively train and evaluate our model. Comparative experiments were conducted to assess performance before and after fine-tuning, as well as against recent CNN-based and Transformer-based vision models.

Our results demonstrate that the pre-trained VLM struggled to accurately detect keypoints out of the box. However, after only two epochs of LoRA fine-tuning, the model's performance significantly improved and reached a comparable level to CNN/Transformer vision base methods. This highlights the potential of leveraging lightweight fine-tuning strategies with VLMs in surgical tool analysis.
% \textbf{Future Work} \

We plan to further optimize the LoRA-based fine-tuning process and explore larger or more specialized VLM architectures. Additionally, we aim to extend the framework to joint estimation of hand-tool interactions and investigate 3D pose estimation for improved performance in dynamic surgical environments.

\textbf{Acknowledgment}: KD is supported by the Royal Thai government scholarship from ministry of higher education, science, research and innovation.

\end{document}